 \documentclass[12pt]{elsarticle}

\usepackage{amssymb}
\usepackage{mathtools}
\usepackage{subfigure}
\usepackage{graphicx}
\usepackage{tabularx,booktabs}
\usepackage{xcolor}
\usepackage{multirow}
\usepackage{lineno}
\usepackage{url}

\journal{arXiv}

\begin{document}

\begin{frontmatter}

	\title{Planogram Compliance Control via Object Detection, Sequence Alignment, and Focused Iterative Search}
	
	\author[erkin]{M. Erkin Y\"{u}cel}
	\ead{mehmety@migros.com.tr}
	
	\author[cem]{Cem \"{U}nsalan}
	\ead{cem.unsalan@marmara.edu.tr}
	
	
	
	\begin{abstract}
	Smart retail stores are becoming the fact of our lives. Several computer vision and sensor based systems are working together to achieve such a complex and automated operation. Besides, the retail sector already has several open and challenging problems which can be solved with the help of pattern recognition and computer vision methods. One important problem to be tackled is the planogram compliance control. In this study, we propose a novel method to solve it. The proposed method is based on object detection, planogram compliance control, and focused and iterative search steps. The object detection step is formed by local feature extraction and implicit shape model formation. The planogram compliance control step is formed by sequence alignment via the modified Needleman-Wunsch algorithm. The  focused and iterative search step aims to improve the performance of the object detection and planogram compliance control steps. We tested all three steps on two different datasets. Based on these tests, we summarize the key findings as well as strengths and weaknesses of the proposed method.
	\end{abstract}
	
	\begin{keyword}
	object detection \sep planogram compliance control \sep sequence alignment \sep focused search
	\end{keyword}
	
	\end{frontmatter}


\section{Introduction}\label{section:Introduction}

Smart retail stores are becoming the fact of our lives. Several computer vision and sensor based systems are working together to achieve such a complex and automated operation. Besides, the retail sector already has several open and challenging problems which can be solved by the help of pattern recognition and computer vision methods. One important problem to be tackled is the planogram compliance control. Planogram is the diagram showing placement of products in shelves of a retail store. Usage of the planogram aims to improve sales by displaying products in a way that inspires the customer to purchase them. The suppliers also pay for the shelves with high selling potential. Or, they apply discounts to their products to be able to place them to these shelves. The planogram also ensures consistency between branches of the same retail store.

\begin{figure}[htbp]
	\centering
	\subfigure[Visual representation of the planogram of the shelf.]{\includegraphics[width=0.9\columnwidth]{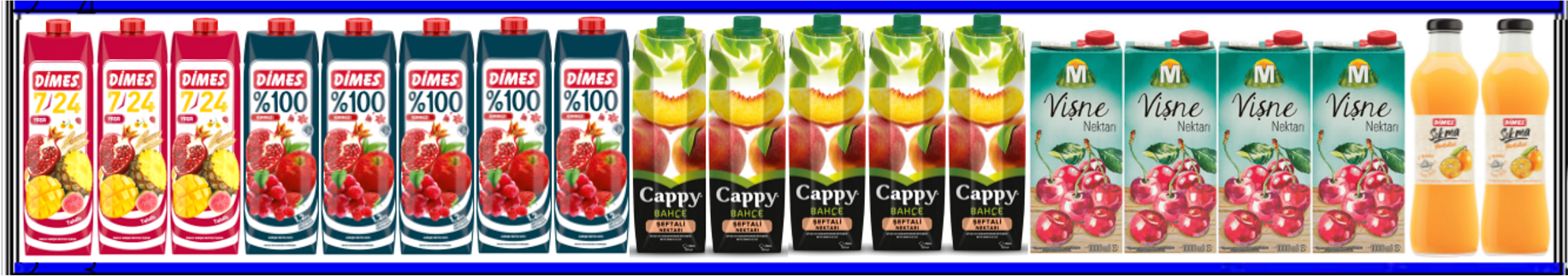}\label{fig:shelf_planogram}}\\
	\subfigure[Fully planogram compliant shelf.]{\includegraphics[width=0.9\columnwidth]{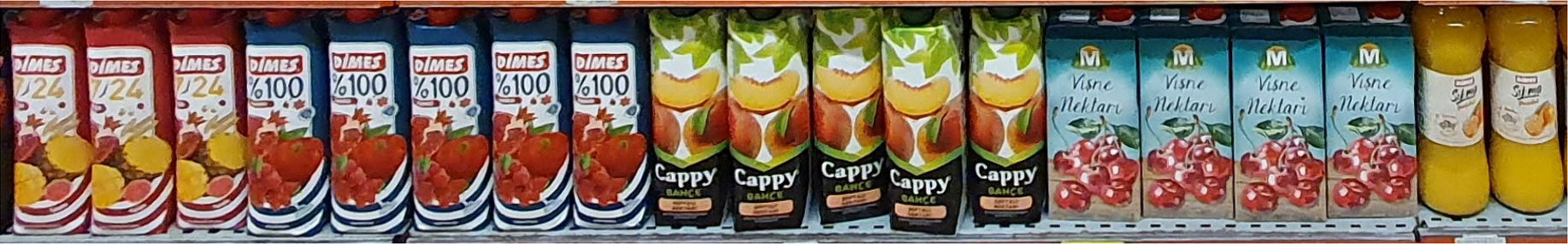}\label{fig:shelf_full}}\\
	\subfigure [Partially planogram compliant shelf.]{\includegraphics[width=0.9\columnwidth]{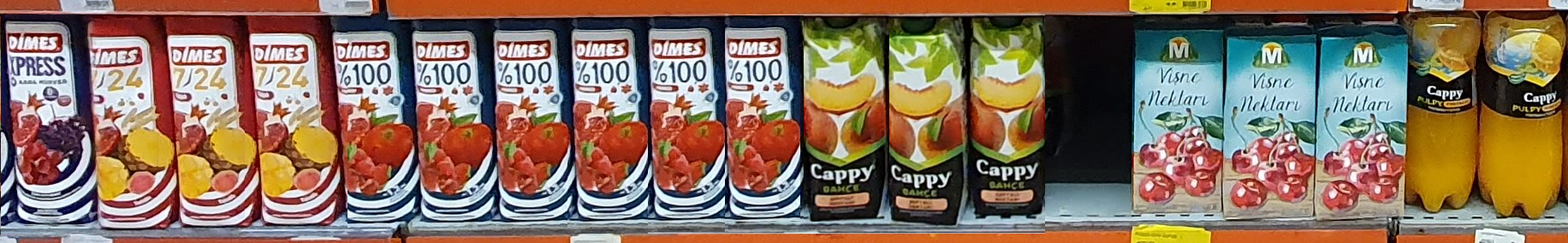}\label{fig:shelf_partial}}\\
	\caption{A sample planogram and corresponding shelf images.}\label{fig:test_images}
\end{figure}

Although planogram usage has such benefits, placement of the products on a shelf may not comply with it. A shelf is considered planogram compliant when all the products on it are in the right place and quantity. However, this may not be satisfied due to several reasons. The products can be misplaced by human error, low on stock or out-of-stock. These situations not only result in loss of sales to the retailer, they also cause the retailer to pay penalty fees to the supplier. According to Shapiro~\cite{Shapiro}, the planogram compliance ratio is only 70\% on average in an average retail store. This results huge sale losses for the retailer. According to the same study, resetting planogram, which is re-allocating all the products, can improve the sales up to 7.8\%  in two weeks. Therefore, retailers must continuously organize the shelves in compliant with the planogram to maximize overall sales. In order to explain the planogram compliance problem further, we provide a sample visual representation of the planogram and corresponding two shelf images in Fig.~\ref{fig:test_images}.

We provide visual representation of the planogram for a shelf in Fig.~\ref{fig:shelf_planogram}. This is the reference planogram which shows how the products should be placed. In this planogram, there are five different products with a total of 19 items in them. Images for these items have been collected from the web such that they have high resolution. We provide a shelf image, taken by camera, fully compliant with the reference planogram in Fig.~\ref{fig:shelf_full}. As can be seen here, although the order and number of the items on the shelf are the same as the reference planogram, objects on the shelf have different viewing angle and non-ideal placement. Besides, there are differences in product images of the planogram and fully compliant shelf in terms of brightness and color. We next provide a shelf image, taken by camera, partially compliant with the reference planogram in Fig.~\ref{fig:shelf_partial}. This image represents most of the planogram compliance problems. We can count them as follows. There is an extra item placed on the leftmost part of the shelf. Hence, all other items are right shifted accordingly. The third product group has one extra item in it. The fourth product group has two less items in it. The fifth product group has one less item in it. There is an empty slot between the fourth and fifth product groups. There is no way we can determine the product this empty slot belongs to. The sixth product group has items resembling actual items, but they are different. Therefore, the partially planogram compliant shelf image in Fig.~\ref{fig:shelf_partial} gives information on the challenges we can face on the planogram compliance control.

Planogram compliance control is conventionally performed manually by an employee in retail stores. The employee walks around aisles and controls shelves once or twice a day for their compliance to the planogram. If the employee detects a problem, then the products on the shelf are re-arranged according to the planogram. However, this method is labor intensive and prone to human error. In an alternative approach, the employee uses a mobile device to take photos of shelves \cite{planorama}. Then, these photos are sent to cloud to check the shelf status. Even though this method eliminates the human error, it is still labor intensive. Another method in the planogram compliance control is using fixed IP cameras instead of mobile devices. Unfortunately, using multiple cameras increases the cost \cite{shelfie}. One recent approach in the planogram compliance control is using robots \cite{simbe,fellowrobots,badger}. In these systems, the mobile robot moves around aisles of the store and collects images and data from the shelves using onboard cameras and sensors. As the robot knows the exact location of itself in the store, it can perform the planogram compliance control. Unfortunately, these systems are fairly expensive.

There are  several methods in literature based on computer vision and pattern recognition to solve the planogram compliance control and object detection from shelf images. Tonioni and Di Stefano~\cite{Tonioni} proposed a method to solve the problem in three steps. First, they detect objects from the shelf image via generalized Hough transform and local feature extraction. Second, they use sub-graph matching to compare the reference planogram and actual item placements. Third, they apply an iterative approach to locate the undetected objects. Our proposed method differs from Tonioni and Di Stefano's work in the second and third steps. Liu~\emph{et al.}~\cite{Liu} provided a planogram compliance control method based on repetitive image patterns. They did not provide an object detection step in their study. There are also related studies focusing on object detection in retail environment. George and Floerkemeier~\cite{George} formed discriminative random forests via all product images. They used SIFT with only one training image per product for image classification. Then, they use deformable dense pixel matching and genetic algorithm optimization. George and Floerkemeier did not provide a planogram compliance control step. Instead, they focused on multi-instance object detection in retail stores. Rosado~\emph{et al.}~\cite{Rosado} used panoramic images to locate stocked out products in shelves without object detection. Jund~\emph{et al.}~\cite{Jund} used CNN to recognize products on the shelf. Unfortunately, CNN needs several images from the same product type for training. Franco~\emph{et al.}~\cite{Franco} first extracted corners from the shelf image. They assumed that each corner is a candidate for being a member of a rectangular object. Then, they formed a window and compared histograms to check whether there is an object there. Afterward, they benefit from CNN for fine selection. Karlinsky~\emph{et al.}~\cite{Karlinsky} first extracted global information such as location of the shelf. Then, they proposed three methods as local feature extraction (via SURF) and implicit shape model usage, histogram of oriented gradients feature extraction and sliding window usage, and local feature extraction (via SURF) and bag of words usage. Their first and third approaches are similar to the object detection step proposed in this study. However, Karlinsky~\emph{et al.} did not consider planogram compliance control in their method. This is the main difference between both studies. Santra~\emph{et al.}~\cite{Santra2} used both object and part level features to detect fine-grained objects in the shelf image. They proposed a reconstruction-classification network to find object level features. Then, they used BRISK to find keypoints on the product image. They encoded the last level of CNN with the convolutional LSTM to describe the part level feature at each keypoint. Finally, they used both object and part level features to detect fine-grained objects. There are two related studies on detecting objects from densely packed images such as acquired from shelves in retail stores. However, the aim is not object recognition in them. The first study is proposed by Goldman~\emph{et al.}~\cite{Goldman} which introduces the SKU100k dataset. It uses different deep learning models on the dataset to detect objects. The second study is proposed by Ye~\emph{et al.}~\cite{Choi} which uses the same dataset on different deep learning models for object detection. There is also a recent review paper by Santra and Mukherjee~\cite{Santra}. The reader can check it for both traditional and deep learning methods for automatic identification of products in retail stores.

In this study, we propose a novel method to detect objects from densely packed shelf images and control planogram compliance. The object detection step depends on local feature extraction and implicit shape model representation. Planogram compliance control step is based on the modified Needleman-Wunsch algorithm which is introduced to align DNA sequences. We also introduce a focused and iterative search step to improve the object detection and planogram compliance control steps. These three steps differ from the ones in literature such that they improve both object detection from shelf images and planogram compliance control sequentially.

The layout of the study is as follows. We first introduce the parts of the object detection step in the proposed method. Therefore, we start with local feature extraction and their representation. Afterward, we explain how brute force search and implicit shape model can be used to detect object center points in a given shelf image. Then, we introduce a bounding box extraction method based on this information. We next explain the planogram compliance control step of the proposed study. To do so, we start with representing the extracted object information as an abstract planogram. Afterward, we introduce the modified Needleman-Wunsch algorithm for planogram compliance control. Then, we explain the focused and iterative search step to improve the performance of the proposed method. We next test the proposed method on two different datasets for the object detection and planogram compliance control steps. Finally, we summarize key findings of the proposed method.

\section{Object Detection and Bounding Box Extraction}\label{sec:objectdetection}

The proposed planogram compliance control method starts with object detection from the shelf image. This operation starts with local feature extraction. Then, we represent the extracted features to be processed in vector form. Afterward, we apply brute force search to match local features extracted from the object of interest and shelf image. This leads to object center point detection via implicit shape model. Finally, we extract bounding boxes for the detected objects on the shelf. We explain all these operations in this section.

\subsection{Local Feature Extraction Methods}\label{sec:localfeatureextraction}

Objects on a retail shelf generally have complex front views. Moreover, there are several nearby objects with almost the same or similar views. Extracting local features and forming an object detection framework using them is more promising compared to using global features. Therefore, we picked five well-known local feature extraction methods as SIFT~\cite{Lowe}, SURF~\cite{Bay}, ORB~\cite{Rublee}, AKAZE~\cite{Alcantarilla}, and BRISK~\cite{Leutenegger}. These methods extract keypoints and a corresponding feature vector from each keypoint as local features via different approaches. We will benefit from the extracted local features and keypoints in detecting objects in the shelf image.

\subsection{Representing the Extracted Local Features in a General Framework}

Since we picked five different local feature extraction methods in this study, we form a general framework to represent them in this section. This will help us to represent the proposed method in a more general framework. Let $I(x, y)$ represent a shelf image with $J$ different objects in it. Assume that the $j^{th}$ object on the shelf has a model (or planogram) image $I_j(x, y)$ for $j = 1, \ldots, J$. Let $w_j$ and $h_j$ be the width and height of $I_j(x, y)$, respectively. We take the model image as tight as possible. Hence, the object has center point $(w_j/2, h_j/2)$.

We can extract local features from $I_j(x, y)$ to represent the $j^{th}$ object. Let $\overrightarrow{k_{jl}}$ be the feature vector extracted at the keypoint $(x_{jl}, y_{jl})$ for $l = 1, \ldots, L_j$ such that $L_j$ is the total number of keypoints extracted using one of the methods given in Section~\ref{sec:localfeatureextraction}. Hence, we will have the merged vector form $\overrightarrow{f_{jl}} = (\overrightarrow{k_{jl}}, x_{jl}, y_{jl})$ for $l = 1, \ldots, L_j$ to represent the $j^{th}$ object with $L_j$ keypoints.

We should also extract local features from the shelf image, $I(x, y)$, to detect objects in it. Based on the previous definitions, we can represent the merged vector for the image as $\overrightarrow{f_m} = (\overrightarrow{k_m}, x_m, y_m)$  for $m = 1, \ldots, M$. Here, $\overrightarrow{k_m}$ is the feature vector extracted at the keypoint $(x_m, y_m)$. $M$ is the total number of extracted keypoints from the shelf image. Moreover, let $w_s$ and $h_s$ be the width and height of $I(x, y)$, respectively. We will use this global information in implicit shape model and bounding box extraction steps in Sections \ref{sec:implicitshapemodel} and \ref{sec:boundingboxextraction}, respectively.

\begin{figure}[htbp]
	\centering
	\subfigure[]{\includegraphics[width=0.06\columnwidth]{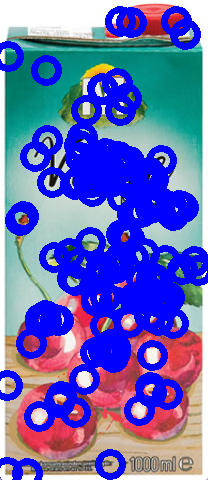}\label{fig:o4_kps}}
	\subfigure[The shelf image.]{\includegraphics[width=0.9\columnwidth]{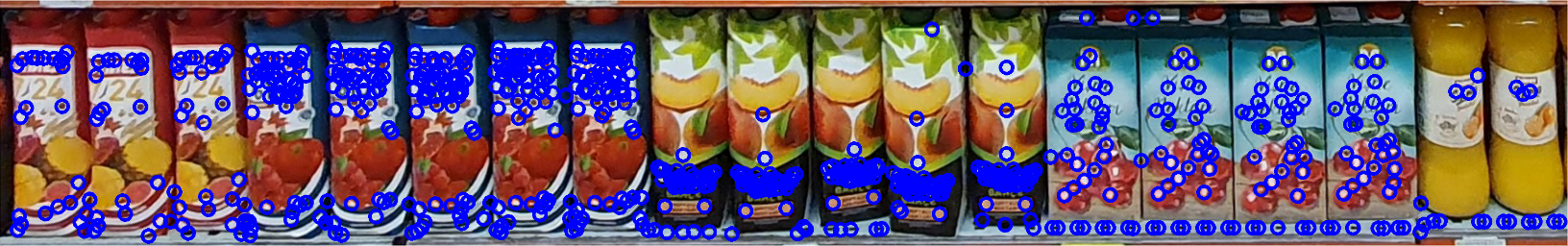}\label{fig:shelf_full_kps}}
	\caption{Extracted keypoints from the fourth object type and shelf images. (a) The object image.}\label{fig:test_images_keypoints}
\end{figure}

We can explain the local feature representation operation on a sample scenario. Let's pick the fourth object type in the planogram and shelf image given in Fig. \ref{fig:shelf_planogram} and \ref{fig:shelf_full}, respectively. We provide most of the extracted SIFT keypoints, as blue circles, in Figs. \ref{fig:o4_kps} and \ref{fig:shelf_full_kps}, respectively. As can be seen in these figures, the extracted keypoints are located mostly on corners and edges in the images. We will use the extracted local feature vectors in these locations for matching next.

\subsection{Brute Force Search for Local Feature Matching}

One way of finding the object of interest in the shelf image is matching extracted local feature vectors from the $j^{th}$ object, $\overrightarrow{k_{jl}}$ for $l = 1, \ldots, L_j$, and shelf image, $\overrightarrow{k_m}$ for $m = 1, \ldots, M$. To do so, we apply brute force search between all combinations and calculate the distance between these two local feature vector groups as $d_{m,jl} = \Vert \overrightarrow{k_m} - \overrightarrow{k_{jl}}\Vert $. Here, we benefit from different distance metrics for different local feature vector types. To be more specific, we use the Euclidean distance for SIFT and SURF since they have float values in their feature vectors. We use the Hamming distance for BRISK, ORB, and AKAZE since they have binary values in their feature vectors.

Lowe~\cite{Lowe} proposed a ratio test to increase the robustness of the SIFT algorithm while finding the best matching local features. The aim of this test is to eliminate features that are not distinct enough even though the distance between them may be below a given threshold. We also use the ratio test to keep only strong matches not only for SIFT, but also for all considered local feature extraction methods. Therefore, we calculate the ratio as

\begin{equation}\label{eqn:loweratio}
	\delta_{m,j} = \frac{d_{m,jl}}{d_{m,jl'}}
\end{equation}

\noindent where $d_{m,jl}$ and $d_{m,jl'}$ denote the smallest and second smallest distance between the feature vector $\overrightarrow{k_m}$ extracted from $I(x,y)$ and the feature vector $\overrightarrow{k_{jl}}$ for $l = 1, \ldots, L_j$ extracted from $I_j(x, y)$.

The idea in using Eqn.~\ref{eqn:loweratio} is that there should be sufficient difference between the best and the second-best matches. Hence, if $\delta_{m,j}$ is smaller than a matching threshold $\tau$, then we assume that $\overrightarrow{k_m}$ and $\overrightarrow{k_{jl}}$ matched. Setting $\tau$ to a high value increases the number of matches. However, this also increases false matches between feature vectors. On the other hand, setting $\tau$ to a low value keeps only strong matches. Hence, some possible matches may be missed.

Lowe suggests setting $\tau=0.8$ gives the best result for SIFT. Instead of such a constant value, we set the matching threshold as

\begin{equation}\label{eqn:matchthrehold}
	\tau_\alpha = 1 - 0.15\alpha
\end{equation}

\noindent where $\alpha$ is the iteration parameter (initially set to one) to be introduced in Section~\ref{sec:focusedanditerativesearch}.

Assume that we have $N_j$ local feature vectors from $I(x,y)$ satisfying the threshold constraint for the $j^{th}$ object. Hence, we will have $\left\{\overrightarrow{f_{jn}}\right\}  \subset \left\{\overrightarrow{f_m}\right\} $ for $n = 1, \ldots, N_j$ and $N_j \leq  M$. Each matching feature vector has a corresponding feature vector from the $j^{th}$ object representation. Assume that the $n^{th}$ feature vector from $I(x,y)$ matches the $b^{th}$ feature vector from $I_j(x,y)$. We can represent this match as $\overrightarrow{f_{jn}} \Leftrightarrow \overrightarrow{f_{jb}}$.

\begin{figure}[htbp]
	\centering
	\subfigure[]{\includegraphics[width=0.06\columnwidth]{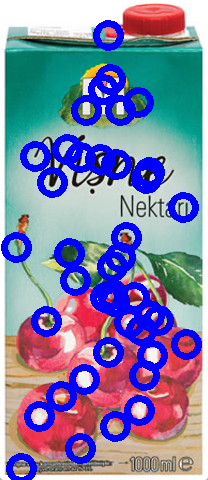}\label{fig:o4_mt}}
	\subfigure[The shelf image.]{\includegraphics[width=0.9\columnwidth]{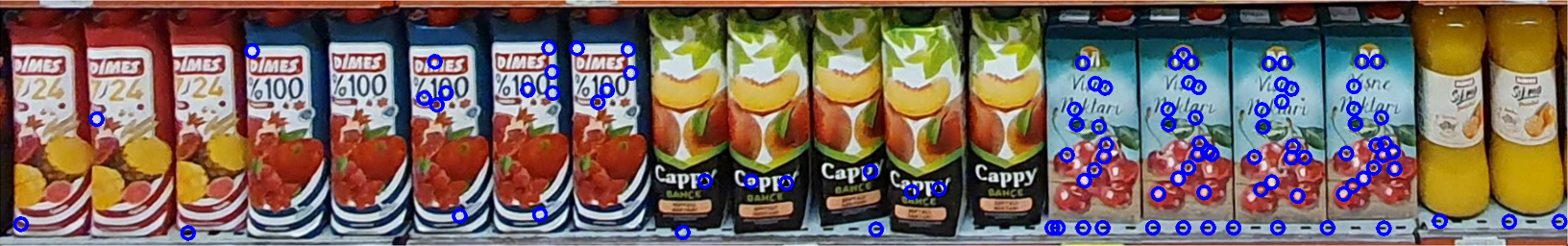}\label{fig:shelf_full_mt}}
	\caption{Matched keypoints between the fourth object type and shelf images. (a) The object image.}\label{fig:test_images_matched}
\end{figure}

We apply the brute force search to the object and shelf images in Fig.~\ref{fig:test_images_keypoints}. We provide the matched keypoints, as blue circles, in Fig.~\ref{fig:test_images_matched}. As can be seen in the figure, the matched keypoints lie on the correct object in the shelf image most of the times. However, there are also false matches in the image. We will handle them in the next section.

\subsection{Implicit Shape Model}\label{sec:implicitshapemodel}

As we match local features, we can use implicit shape model (ISM) to detect the center of matched objects in the shelf image $I(x,y)$ \cite{Leibe}. We picked ISM for this operation since it allows detecting multiple objects of the same type in the image. We modified the original ISM to serve our purposes better. Our modification is in the matching phase of the method. Leibe~\emph{et al.} uses agglomerative clustering to search for matching image patches in their original implementation. Agglomerative clustering requires multiple training objects to group similar image patches and clustering their features. Besides, it has a high computation cost. We have only one representative image for each object in our problem. Therefore, we use only local features and keypoints to create a voting matrix in ISM. This allows us to find object centers in the shelf image fairly fast.

To implement the ISM suitable for our problem, we first create an empty voting matrix $V_j(x, y)$ for the $j^{th}$ object for $j = 1, \ldots, J$. Hence, we form $J$ distinct voting matrices for $J$ candidate objects to be detected in the shelf image, $I(x,y)$. Each voting matrix has the same size as with the shelf image.

Matching $\overrightarrow{f_{jn}} \Leftrightarrow \overrightarrow{f_{jb}}$ can be taken as evidence for one or more objects' presence in the shelf image. Therefore, each $\overrightarrow{f_{jn}}$ votes for its candidate object center location. For the $n^{th}$ matched local feature of the $j^{th}$ object, we will have the voting coordinate as

\begin{equation}\label{eqn:votingcoordinates}
	\begin{split}
		\hat{x}_n &= x_{n} + \beta_j r_n \cos (\Theta_n) \\
		\hat{y}_n &= y_{n} + \beta_j r_n \sin (\Theta_n)
	\end{split}
\end{equation}

\noindent where

\begin{equation}\label{eqn:votingcoordinates2}
	\begin{split}
		r_n &= \left[(w_j/2 - x_{jb})^2 + (h_j/2 - y_{jb})^2\right] ^{1/2} \\
		\Theta_n &= \arctan\left(\frac{h_j/2 - y_{jb}}{w_j/2 - x_{jb}}\right)
	\end{split}
\end{equation}

\noindent Here $\beta_j = h_s / h_j$, $h_s$ being the height of $I(x, y)$.

Based on these definitions, we can form the voting matrix for the $j^{th}$ object as

\begin{equation}\label{eqn:votingmatrix}
	V_j(x,y) = \sum_{n = 1}^{N_j} \gamma_n  \exp \left[-\frac{(x-\hat{x}_n)^2 + (y-\hat{y}_n)^2}{2\sigma^2} \right]
\end{equation}

\noindent where $\gamma_n = 1 - d'_{n,jb}$, $d'_{n,jb}$ being the min-max normalized form of $d_{n,jb}$ \cite{Han}. $\sigma$ is the standard deviation of the Gaussian kernel used as the voting function. We set $\sigma=7$ based on the average size of the objects to be detected. Since we form the voting matrix for each object type separately, it represents vote values for that object type only.

\begin{figure}[htbp]
	\centering	
	\includegraphics[width=0.9\columnwidth]{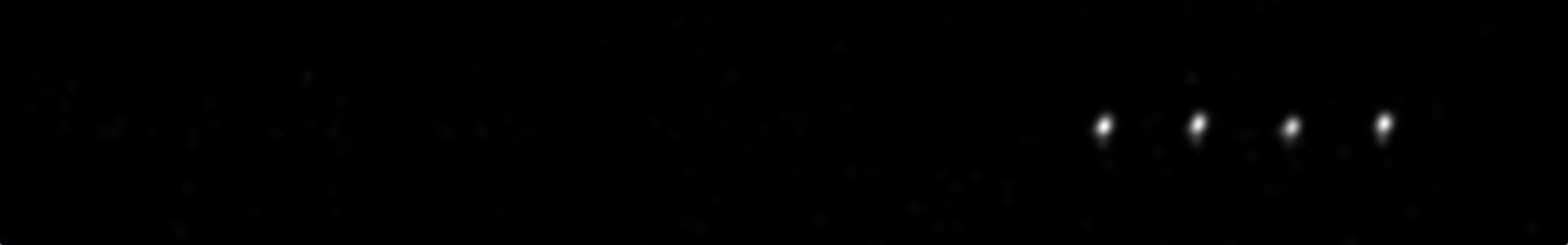}
	\caption{Voting matrix, $V_j(x,y)$, formed for the fourth object type in the shelf image.}\label{fig:votingmatrix}
\end{figure}

We next consider the matched features given in Fig.~\ref{fig:test_images_matched}. We provide the voting matrix, formed for the fourth object type in Fig.~\ref{fig:o4_mt}, as in Fig.~\ref{fig:votingmatrix}. In this figure, bright locations indicate high votes. As can be seen in this figure, the votes cumulate around possible center locations of the fourth object type in the shelf image.

We can hypothesize that modes of $V_j(x,y)$ are possible object centers for the $j^{th}$ object in the shelf image. However, all modes do not correspond to an object center. Therefore, we take a location as a valid object center when the cumulate votes there exceed a threshold value. To do so, we define a dynamic threshold as

\begin{equation}\label{eqn:probabilitythreshold}
	\tau_v = \frac{\alpha}{2} \max \left(V_j(x,y)\right)
\end{equation}

\noindent where $\alpha$ is the iteration parameter (initially set to one) to be introduced in Section~\ref{sec:focusedanditerativesearch}. As a result, we obtain the candidate object centers from $V_j(x,y)$ as

\begin{equation}\label{eqn:objectcenters}
	(x_{jc},y_{jc}) = \arg\max_{(x,y)} \left( V_j(x,y)\right)
\end{equation}

\noindent such that $V_j(x_{jc},y_{jc}) > \tau_v$.

Assume that we obtain $C_j$ candidate object centers for the $j^{th}$ object. Hence, we will have $(x_{jc},y_{jc})$ for $c = 1, \ldots, C_j$ and $j = 1, \ldots, J$ and their vote values as $V_j(x_{jc},y_{jc})$. We will use these in extracting bounding boxes for the objects in the shelf image in the next section.

\begin{figure}[htbp]
	\centering	
	\includegraphics[width=0.9\columnwidth]{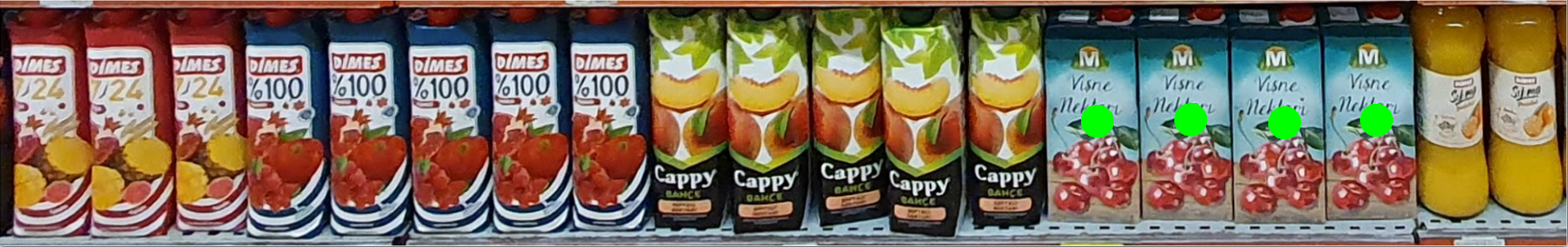}
	\caption{Extracted object centers for the fourth object type in the shelf image.}\label{fig:detectedobjectcenters}
\end{figure}

We apply the candidate object center extraction procedure to the voting matrix in Fig.~\ref{fig:votingmatrix}. As a reminder, this voting matrix has been formed for the fourth object type, given in Fig.~\ref{fig:o4_mt}, in the shelf image. We provide the extracted object centers, as green dots, in Fig.~\ref{fig:detectedobjectcenters}. As can be seen in this figure, the fourth object has been successfully located by its center coordinates in the shelf image.

\subsection{Bounding Box Extraction}\label{sec:boundingboxextraction}

As we detect possible object center locations in the shelf image, the next step is forming a bounding box for each location. We have the bounding box representation for the object in the planogram image. We know the ratio of planogram object size wrt the shelf image. Hence, we can form the bounding box for the detected object in the shelf image directly. In this implementation, the detected object in the shelf may not have a rectangular shape. It may be shifted or rotated. Therefore, the object in the shelf image may be taken as smaller or larger than the planogram image. To handle this issue, we add a 20\% overlap tolerance in all operations.

Let's formalize the bounding box extraction operation based on the previous definitions. We should form a bounding box for each object center, $(x_{jc},y_{jc})$ for $c = 1, \ldots, C_j$ and $j = 1, \ldots, J$. To do so, we can use the global information of $I(x, y)$ and $I_j(x, y)$ to fit a bounding box around the object center. We represent a bounding box by its top-left and bottom-right coordinates. Since the $c^{th}$ object center is associated with the $j^{th}$ object, we will have

\begin{equation}\label{eqn:boundingboxcoordinates}
	B_{jc} = \left[ (x_{jc} - \frac{\beta_j w_j}{2}, y_{jc} - \frac{\beta_j h_j}{2}), (x_{jc} + \frac{\beta_j w_j}{2}, y_{jc} + \frac{\beta_j h_j}{2}) \right]
\end{equation}

\noindent where $\beta_j$ is the normalizing factor introduced in Eqn.~\ref{eqn:votingcoordinates2}. At this point, we have bounding box, $B_{jc}$, and vote value, $V(x_{jc},y_{jc})$, for the candidate object center $(x_{jc},y_{jc})$.

Some bounding boxes may overlap. This means that there are more than one candidate object in the given location. We should eliminate weak candidates. To do so, we check the intersection over union (IoU) of all bounding boxes \cite{Rezatofighi}. If the IoU value of the two bounding boxes is above an overlap threshold, experimentally set as 20\%, then we discard the candidate object center with a lower vote value. Hence, we perform non-maxima suppression. At the end of this operation, we detect a total of $D$ objects, $D \leq \sum_{j = 1}^{J} C_{j}$, from the shelf image $I(x, y)$.

Up to now, we explained how to form the bounding box for detected objects in the shelf image. There may also be empty spaces or undetected (unknown) objects in the image. We should analyze these locations as well. Therefore, we next search for empty spaces in the shelf image after detecting all bounding boxes corresponding to the detected objects. To do so, we model the empty space as a black box since it will be dark compared to its surrounding. Then, we apply template matching. We then merge the detected neighboring empty space blocks. As for discriminating the empty space blocks and undetected objects, we first obtain the average width of detected objects in the shelf image. Here, we assume that the undetected object will have similar width as its neighboring objects. If the merged empty space block width is the same or larger than this width, then we label that region as empty space. If we neither detect the object nor an empty space in a part of the shelf image, then we label that region as unknown.

\begin{figure}[htbp]
	\centering
	\subfigure[Fully planogram compliant shelf image.]{\includegraphics[width=0.9\columnwidth]{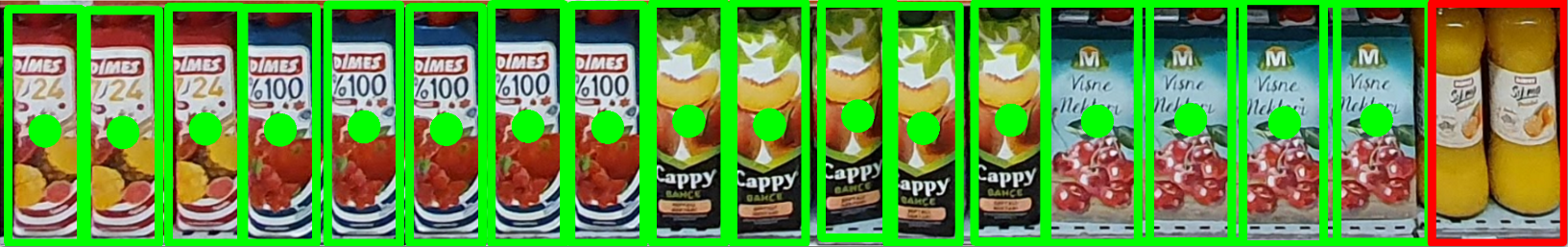}\label{fig:shelf_full_boundingboxes}}\\
	\subfigure[Partially planogram compliant shelf image.]{\includegraphics[width=0.9\columnwidth]{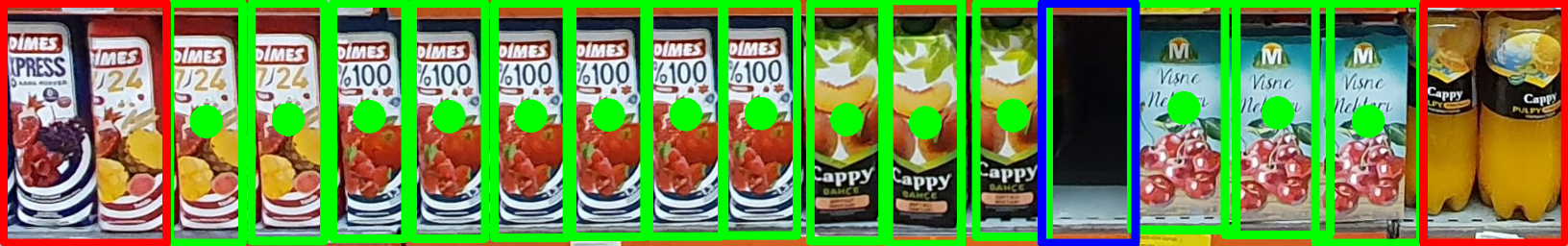}\label{fig:shelf_partial_boundingboxes}}\\
	\caption{Formed bounding boxes for the fully and partially planogram compliant shelf images.}\label{fig:detected_bounding_boxes_1st_iteration}
\end{figure}

We next consider the fully and partially planogram compliant shelf images in Figs. \ref{fig:shelf_full} and \ref{fig:shelf_partial}. We provide the corresponding bounding box detection results in Figs. \ref{fig:shelf_full_boundingboxes} and \ref{fig:shelf_partial_boundingboxes}, respectively. We set a green bounding box for each detected object center in these images. There may be small overlapping areas in these bounding boxes. Since they did not exceed the overlap threshold, they did not pose any problems. There were undetected objects both in the fully and partially planogram compliant images. We take them as unknown and plot the corresponding bounding boxes as red colored.  We will correct these errors by an iterative approach in the following section. There is also an empty space in the partially planogram compliant shelf image. We plot the corresponding bounding box as blue.

\section{Planogram Compliance Control via Sequence Alignment and Focused Iterative Search}

Object detection results in Section~\ref{sec:objectdetection} lead to planogram compliance control. To do so, we first represent the detection results in an abstract planogram format. Then, we propose an iterative search method to focus on undetected or empty object locations in the shelf image. We explain these steps in detail next.

\subsection{Planogram Formation from the Extracted Object Information}

We can form an abstract planogram representation from object detection results. In our representation, there will be spatially sorted object types, number of objects for each type, and bounding box for each object type. Therefore, we should first sort detected objects in the shelf image. We can perform sorting by using horizontal coordinate of the center point, $x_{jc}$, for the bounding box of detected objects.

The proposed planogram compliance control method can handle objects placed on top of each other. The hard constraint from planogram formation for such cases is that these objects should be of the same type. Therefore, we can sort them. There may be two scenarios for the objects placed on top of each other. First, the horizontal coordinates of these objects may be different. Then, they will be sorted as such. Second, if the horizontal coordinates of overlapping two objects are the same, then the one detected first is placed to the left of the second one.

There is also another hard constraint for planogram formation. If there are objects of the same type on a shelf, then they should be placed side by side. Therefore, we next group sorted objects of the same type and standing side by side (or on top of each other) on the shelf. This can be done by checking bounding boxes. As a result, we form a merged bounding box for the grouped objects.

Based on the performed operations in the previous paragraphs, we can represent the detected objects in the shelf image (in sorted and grouped form) as $L_s = \left[ o_d, q_d, B_d\right]$ for $d = 1, \ldots, E$ where $E \leq D$. Here, $o_d$ stands for the detected object group type. $q_d$ stands for the number of objects in the object group $o_d$. $B_d$ stands for the merged bounding box of the grouped objects of type $o_d$. We can call $L_s$ as the detected planogram representation for the shelf.

There should also be the reference (correct or desired) planogram for the shelf. Based on the previous definitions, we can represent it as $L_r = \left[ o_t, q_t, B_t\right]$ for $t = 1, \ldots, T$. Here, $o_t$, $q_t$, and $B_t$ stand for the group type, number of objects in the group, and bounding box, respectively.

We can form the abstract representation of the sample planogram in Fig.~\ref{fig:shelf_planogram}. To do so, we should have executed all the steps up to this point. Based on these, the abstract representation for the reference planogram, $L_r$, will be as in Table~\ref{table:shelf_a_pla}.

\begin{table}[htbp]
	\centering
	\caption{Reference planogram $L_r$ for the shelf.}	\label{table:shelf_a_pla}
	\begin{tabular}{lccccc}
		\toprule
		$o_t$	& $o_1$	& $o_2$	& $o_3$	& $o_4$	& $o_5$	\\
		$q_t$	& 3		& 5		& 5		& 4		& 2		\\
		\bottomrule
	\end{tabular}
\end{table}

We can also form the detected planograms for the fully and partially planogram compliant shelf images in Figs. \ref{fig:shelf_full} and \ref{fig:shelf_partial}. We provide them in Tables \ref{table:shelf_full_pla} and \ref{table:shelf_partial_pla}, respectively. In these tables, 'U' and 'E' represent the undetected (unknown) object and empty space, respectively. To note here, both planograms are formed before applying any iterative improvements, which is the main topic for the next section. Therefore, we will take these as initial formed planograms.

\begin{table}[htbp]
	\centering
	\caption{The planogram $L_s$ for the fully planogram compliant shelf.}\label{table:shelf_full_pla}
	\begin{tabular}{lccccc}
		\toprule
		\textbf{$o_d$}	& $o_1$	& $o_2$	& $o_3$	& $o_4$	& U	\\
		\textbf{$q_d$}	& 3		& 5		& 5		& 4		& 1			\\
		\bottomrule
	\end{tabular}
\end{table}

\begin{table}[htbp]
	\centering
	\caption{The planogram $L_s$ for the partially planogram compliant shelf.}\label{table:shelf_partial_pla}
	\begin{tabular}{lccccccc}
		\toprule
		$o_d$	& U	& $o_1$	& $o_2$	& $o_3$	& E	& $o_4$	& U	\\
		$q_d$	& 1			& 2		& 6		& 3		& 1		& 3		& 1			\\
		\bottomrule
	\end{tabular}
\end{table}

\subsection{The Needleman-Wunsch Algorithm}

We can control planogram compliance by comparing the reference, $L_r$, and detected, $L_s$, planograms for a given shelf. Unfortunately, direct comparison does not work since there can be unknown, extra or missing objects on the shelf. Moreover, we may not know their quantity and location. Therefore, we propose using the Needleman-Wunsch (NW)~\cite{Needleman} algorithm introduced for aligning two DNA sequences. We modify it for our problem to be explained in detail next.

The NW algorithm uses dynamic programming to align two sequences by using a score matrix. In it, columns and rows represent the two DNA sequences to be aligned. The NW algorithm systematically fills the matrix using a substitution score and gap penalty. In other words, the substitution score and gap penalty are used to calculate the score value to be added to the score matrix for the compared element pair of two sequences. Hence, the score matrix keeps the score obtained by comparing all the elements of two sequences with each other.

The NW algorithm has three steps. We modified them to be suitable for our problem as follows. Let $L_s$ and $L_r$ be the two planograms to be aligned. Let $F$ be the score matrix with size $(E+1)\times(T+1)$. Here, $L_s$ and $L_r$ are placed horizontally and vertically to $F$, respectively. For our planogram compliance control problem, we can consider four possibilities for each element pair of the compared two sequences as match, mismatch/substitute, insert, and delete. Let $g_{ins}$ and $g_{del}$ denote the gap penalties for insert and delete, and $s(o_d, o_t)$ the substitution score for match and mismatch/substitute. In the original NW algorithm, these have constant values. In this study, we set the gap penalties and substitution score dynamically based on their importance as $g_{ins} = q_d$, $g_{del} = q_t$, and

 \begin{equation}\label{eqn:subsscore}
 	s(o_d, o_t)  = \begin{cases}
 		+q_t, o_d \Leftrightarrow o_t \\
 		-q_t, o_d \nLeftrightarrow o_t
 	\end{cases}
 \end{equation}

We perform this dynamic valuation based on the required and detected number of items on the shelf. Therefore, we increase the importance of the detected and undetected object numbers. Hence, non-aligned one and ten objects will have different values. We had to define two different gap penalty values for the insert and delete operations. The gap penalty for the delete operation is set as the required number of objects in the reference planogram since we could not detect them. The gap penalty for the insert operation is set as the number of extra items in the detected planogram. The aim here is that the more inserted such false objects are, the more penalty there should be.

Based on the modifications applied to the NW algorithm for the planogram compliance control problem, the first step in implementation is initializing the score matrix $F$. Hence, we should set the first row and column of $F$ as

\begin{equation}\label{eqn:scorematinit}
	\begin{split}
		F(0,0) &= 0\\
		F(0,t) &= F(0,t-1) - 1\\
		F(d,0) &= F(d-1,0) - 1
	\end{split}
\end{equation}

\noindent for $d = 1, \ldots, E$ and $t = 1, \ldots, T$. To note here, we provide the first row and column of the $F$ matrix manually. The initial value $F(0,0)$ should be set as 0. The first row and columns should have values as -1, -2, -3 and so on.

The second step in the modified NW algorithm is the definition of iterations through $F(d,t)$. The original algorithm defines the score for each matrix entry based on the upper, left, and diagonally up-left adjacent entries. Then, the maximum of them is assigned to the current matrix entry as the score. Hence, we can fill the matrix entries for our problem using the recursive formulas

\begin{equation}\label{eqn:scorematrix}
	F(d,t) = \max \begin{cases}
		F(d - 1, t) - g_{ins}, \\
		F(d, t - 1) - g_{del}, \\
		F(d - 1, t - 1) + s(o_d, o_t)
	\end{cases}
\end{equation}

\noindent The recursion ends when the calculations reach the bottom right corner of the score matrix. In other words, the recursion ends when $d = E$ and $t = T$.

The third step in the modified NW algorithm is tracing back the filled matrix entries. Here, we trace back the decisions we have made while calculating each matrix entry. If $F(d,t)$ is calculated from its left adjacent neighbor $F(d - 1, t)$, then there is a delete from $o_d$ according to $o_t$. We denote this by inserting 'D' to $o_d$ and 0 to $q_d$. If $F(d,t)$ is calculated from its upper adjacent neighbor $F(d, t - 1)$, then there is an insert to $o_d$ according to $o_t$. We denote this by inserting 'A' to $o_t$ and 0 to $q_t$. Finally, if $F(d,t)$ is calculated from its up-left adjacent neighbor $F(d - 1, t - 1)$, then there is a match or mismatch/substitute between $o_d$ and $o_t$. Hence, we can move to the next matrix entry. The trace back ends when we reach the upper left corner of the score matrix. In other words, trace back ends when $d = 0$ and $t = 0$.

Now we have aligned object groups as $\hat{o}_d$ and $\hat{o}_t$ and the corresponding object numbers $\hat{q}_d$ and $\hat{q}_t$ are at hand. We can directly compare them to locate the correctly matched items (MT), missing items at correct location (MI), extra items at correct location (ME), and extra or falsely placed items, or empty spaces (NM) in $L_r$ and $L_s$. If $\hat{o}_d \Leftrightarrow \hat{o}_t$, we can have MT, MI, or ME. Then, we can compare $\hat{q}_d$ and $\hat{q}_t$. Here, if $\hat{q}_d = \hat{q}_t$, then we will have MT. If $\hat{q}_d < \hat{q}_t$, then we will have MI. If $\hat{q}_d > \hat{q}_t$, then we will have ME. If $\hat{o}_d \nLeftrightarrow \hat{o}_t$, then we will have NM. Here, the quantity comparison is not necessary. Based on these scenarios, we can define the planogram match ratio between $L_r$ and $L_s$ as

\begin{equation}\label{eqn:matchratio}
	\mu_s = \frac{\sum_{d = 1}^{E} \hat{q}_d \delta_{dt} }{\sum_{t = 1}^{T} \hat{q}_t}
\end{equation}

\noindent where

\begin{equation}\label{eqn:matchratio2}
	\delta_{dt}  =
	\begin{cases}
		1, & \hat{o}_d \Leftrightarrow \hat{o}_t\\
		0, & \hat{o}_d \nLeftrightarrow \hat{o}_t
	\end{cases}
\end{equation}

In Eqn.~\ref{eqn:matchratio}, $\mu_s$ can take values between 0 and 1. When $\mu_s = 0$ this indicates that all of the items on the shelf are falsely placed compared to the reference planogram. When $\mu_s = 1$, this indicates that all items on the shelf are placed in the right place and in correct quantity. In other words, the shelf is fully compliant with the reference planogram. When $0 < \mu_s < 1$, this indicates that the shelf is partially compliant with the reference planogram and some items are either missing, extra or incorrectly placed.

\begin{figure}[htbp]
	\centering
	\includegraphics[width=0.5\columnwidth]{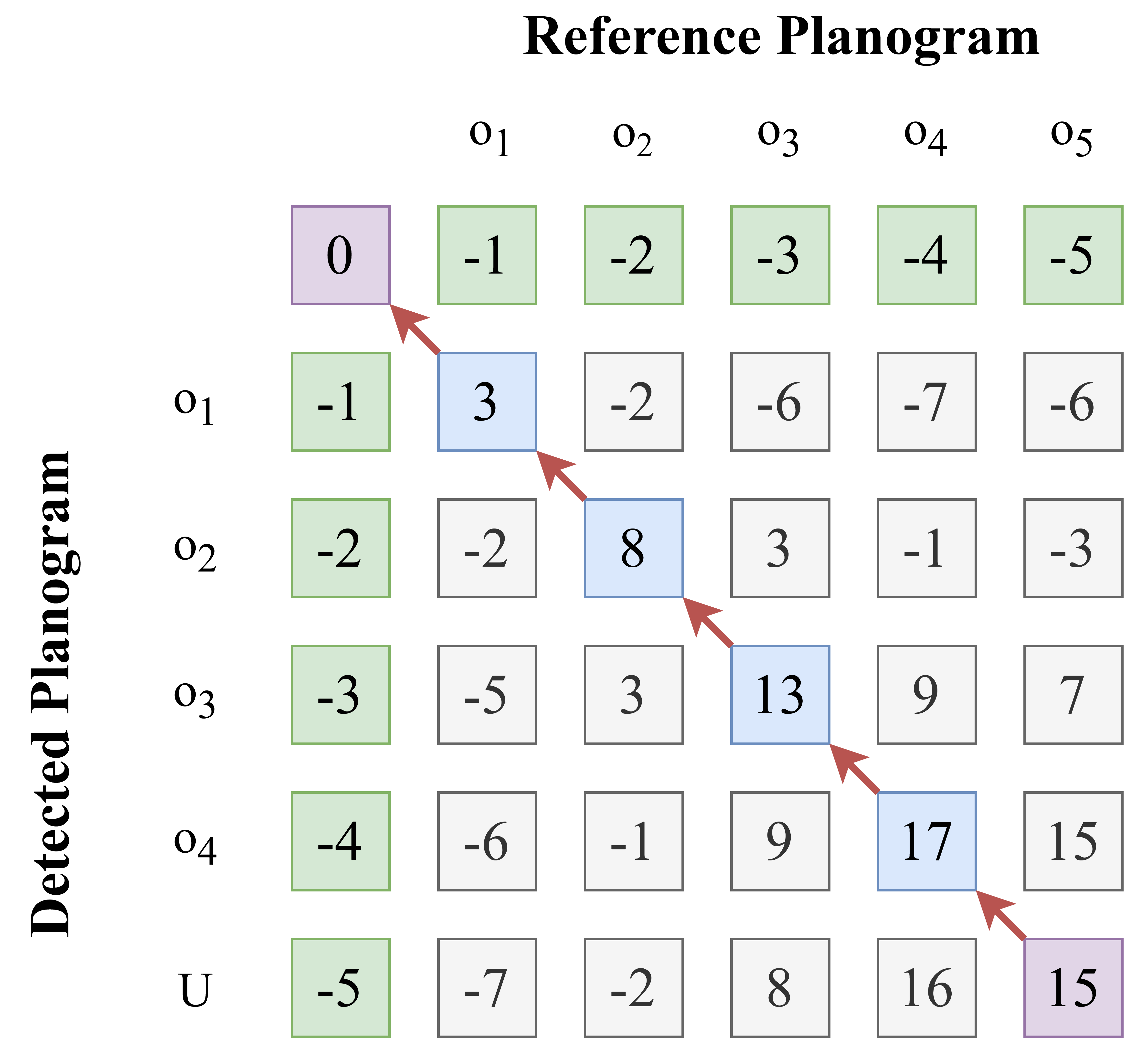}
	\caption{The NW score matrix for the reference and fully compliant shelf planogram alignment.}\label{fig:scorematrix_full}
\end{figure}

We can explain the score matrix formation using the reference and fully planogram compliant shelf planograms given in Tables \ref{table:shelf_a_pla} and \ref{table:shelf_full_pla}. The formed score matrix based on these will be as in Fig.~\ref{fig:scorematrix_full}. We can find the alignment of the reference and detected planograms by following the red arrows in the figure.

Based on Fig.~\ref{fig:scorematrix_full}, the alignment result will be as in Table~\ref{table:shelf_full_align_1st_iteration}. In this table, GT stands for the ground truth. When we compare the aligned planograms, we obtain $\mu=0.89$. This value is not equal to 1 (which is the desired case) since there are still objects in the shelf image we would like to detect. We will handle them in the next section.

\begin{table}[htpb]
	\centering
	\caption{Alignment result of the reference and detected planograms for the fully compliant shelf.}\label{table:shelf_full_align_1st_iteration}
	\begin{tabular}{lccccc}
		\toprule
		$\hat{o}_t$	& $o_1$	& $o_2$	& $o_3$	& $o_4$	& $o_5$		\\
		$\hat{q}_t$	& 3		& 5		& 5		& 4		& 2			\\
		\midrule
		$\hat{o}_d$			& $o_1$	& $o_2$	& $o_3$	& $o_4$	& U			\\
		$\hat{q}_d$			& 3		& 5		& 5		& 4		& 1			\\
		\midrule
		\textbf{Result}			& MT	& MT	& MT	& MT	& NM		\\
		\textbf{GT}				& MT	& MT	& MT	& MT	& MT		\\
		\bottomrule
	\end{tabular}
\end{table}

We can repeat the score matrix formation example using the reference and partially planogram compliant shelf planograms given in Tables \ref{table:shelf_a_pla} and \ref{table:shelf_partial_pla}. Again, we can find the alignment of the reference and detected planograms by following the red arrows in the figure.

\begin{figure}[htbp]
	\centering
	\includegraphics[width=0.5\columnwidth]{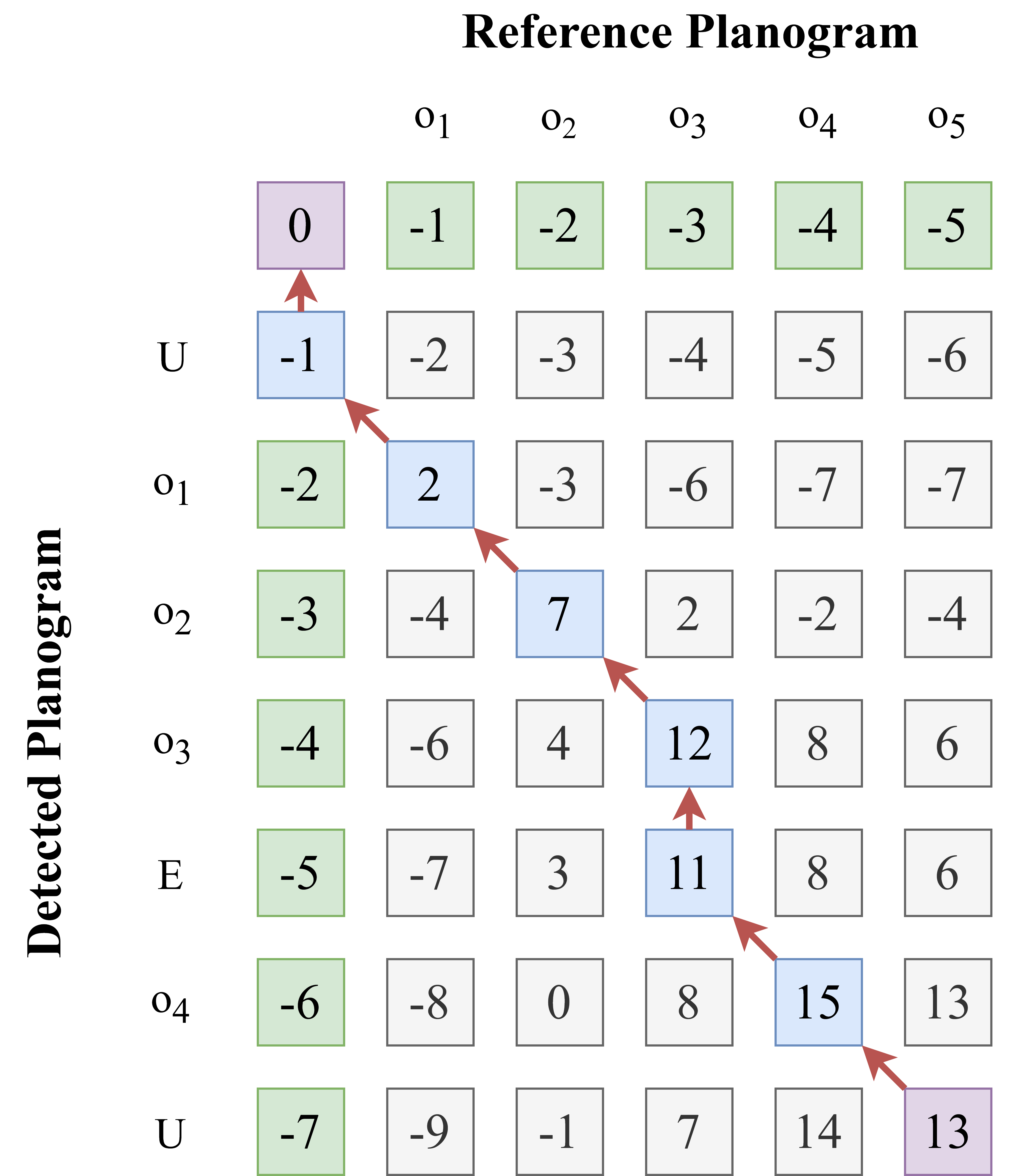}
	\caption{The NW score matrix for the reference and partially compliant shelf planogram alignment.}\label{fig:scorematrix_partial}
\end{figure}

Based on Fig.~\ref{fig:scorematrix_partial}, the alignment result will be as in Table~\ref{table:shelf_partial_align_1st_iteration}. In this table, GT stands for the ground truth. When we compare the aligned planograms, we obtain $\mu=0.68$. As in the previous example, $\mu$ is not equal to the expected value 0.73 since there are still objects in the shelf image we would like to detect. We will handle them in the next section.

\begin{table}[htpb]
	\centering
	\caption{Alignment result of the reference and detected planograms for the partially compliant shelf.}\label{table:shelf_partial_align_1st_iteration}
	\begin{tabular}{lccccccc}
		\toprule
		$\hat{o}_t$	& A		& $o_1$	& $o_2$	& $o_3$	& A		& $o_4$	& $o_5$ 	\\
		$\hat{q}_t$	& 0		& 3		& 5		& 5		& 0		& 4		& 2			\\
		\midrule	
		$\hat{o}_d$			& U		& $o_1$	& $o_2$	& $o_3$	& E		& $o_4$	& U			\\
		$\hat{q}_d$			& 1		& 2		& 6		& 3		& 1		& 3		& 1			\\
		\midrule
		\textbf{Result}			& NM	& MI	& ME	& MI	& NM	& MI	& NM		\\
		\textbf{GT}				& NM	& MT	& ME	& MI	& NM	& MI	& NM		\\
		\bottomrule
	\end{tabular}
\end{table}

\subsection{Focused and Iterative Search}\label{sec:focusedanditerativesearch}

The object detection method and modified NW algorithm introduced in the previous sections may not produce exact planogram alignment result in one iteration. In other words, some of the objects may not be detected or falsely detected in the planogram in one iteration. This is the case most of the times for the planogram compliance control problem since we have only one training image per object. Therefore, we propose a focused and iterative search method to improve the results.

The proposed focused and iterative search method works as follows. We repeat the object detection operation by relaxing the detection constraints and focusing on the region of interest (ROI) in the shelf image $I(x,y)$. We define ROI based on the falsely detected, partially detected or undetected object group locations by the object detection and NW algorithm steps. In other words, we discard the bounding box of correctly matched regions from $I(x,y)$ and focus on the remaining regions as ROI.

As we form the ROI(s), we iteratively decrease the value of $\alpha$ as $\alpha_{new} = 0.75 \alpha_{old}$ in Eqns.~\ref{eqn:probabilitythreshold} and~\ref{eqn:matchthrehold}. As we decrease $\alpha$ in an iteration, the number of matched features increases. We decrease $\alpha$ by multiplication to have a fast decrement in initial iterations. We picked the multiplier 0.75 such that we can have at most 10 iterations for obtaining a reasonable result.

We detect new objects based on the new $\alpha$ value at each iteration. Then, we update $L_s$ based on the previous (existing) and new object detection results. Afterward, we apply the NW algorithm to the updated $L_s$ and $L_r$. We finally calculate $\mu$ for this iteration. We repeat this operation until $\mu = 1$ or $\mu$ does not change for six successive iterations. As the iterations end, we have the planogram compliance control result at hand.

\begin{figure}[htbp]
	\centering
	\subfigure[Fully planogram compliant shelf image.]{\includegraphics[width=0.9\columnwidth]{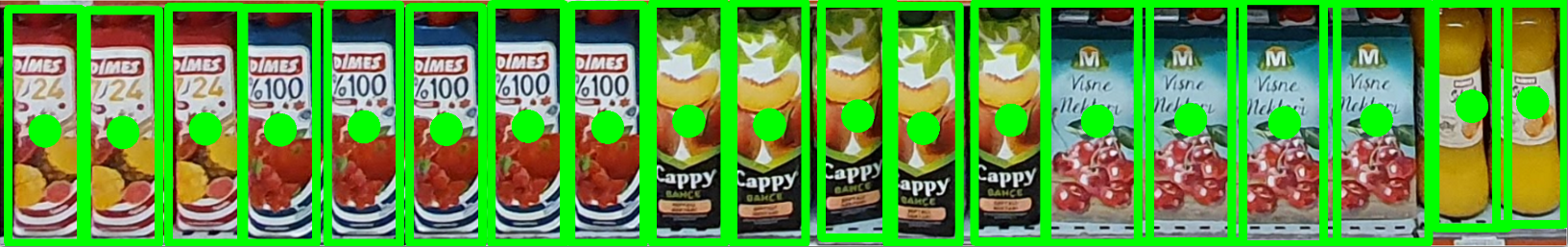}\label{fig:shelf_full_boundingboxes_iteration}}\\
	\subfigure[Partially planogram compliant shelf image.]{\includegraphics[width=0.9\columnwidth]{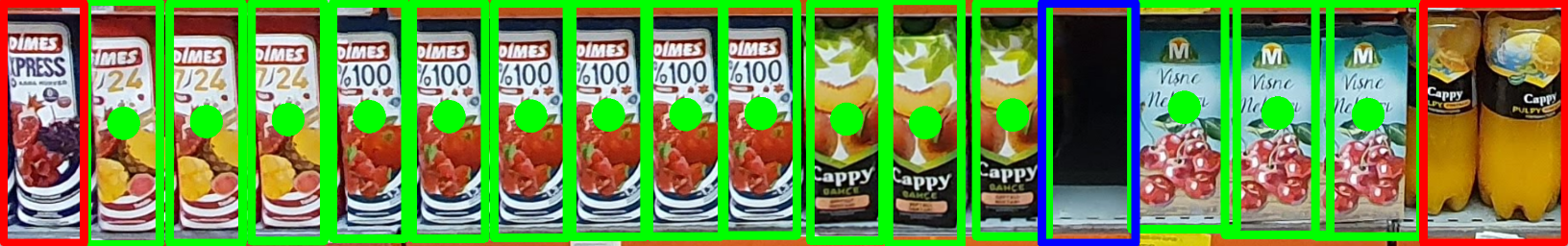}\label{fig:shelf_partial_boundingboxes_iteration}}\\
	\caption{Bounding box of the detected objects in the shelf images with focused and iterative search.}\label{fig:detectedboundingboxes}
\end{figure}

We apply the focused and iterative search method to our fully and partially planogram compliant shelf control examples in the previous section. We provide the final detected objects and corresponding bounding boxes for them in Fig.~\ref{fig:detectedboundingboxes}. As can be seen in Fig.~\ref{fig:shelf_full_boundingboxes_iteration}, we can detect the initially undetected two rightmost objects on the right side of the fully planogram compliant shelf after the focused and iterative search step. Hence, the $\mu$ value becomes 1 (as expected) for this case. As for the partially planogram compliant shelf image, we provide the result in Fig.~\ref{fig:shelf_partial_boundingboxes_iteration}. As can be seen in this image, we can detect the undetected leftmost object and missing second object in the shelf image after the focused and iterative search step. Hence, the $\mu$ value increases to 0.73 for this case.

We provide the final obtained planogram for the fully compliant shelf in Table~\ref{table:plaalignfully}. As can be seen in this table, the focused and iterative search method led to detecting all objects in the shelf image. Hence, the alignment result became the same as the ground truth (GT) as expected.

\begin{table}[htpb]
		\centering
		\caption{Alignment result of the reference and detected planograms for the fully compliant shelf.}\label{table:plaalignfully}
		\begin{tabular}{lccccc}
			\toprule
			$\hat{o}_t$	& $o_1$	& $o_2$	& $o_3$	& $o_4$	& $o_5$\\
			$\hat{q}_t$	& 3		& 5		& 5		& 4		& 2			\\
			\midrule
			$\hat{o}_d$			& $o_1$	& $o_2$	& $o_3$	& $o_4$	& $o_5$	\\
			$\hat{q}_d$			& 3		& 5		& 5		& 4		& 1			\\
			\midrule
			\textbf{Result}			& MT	& MT	& MT	& MT	& MT		\\
			\textbf{GT}				& MT	& MT	& MT	& MT	& MT		\\
			\bottomrule
		\end{tabular}
	\end{table}

We next provide the final planogram obtained for the partially compliant shelf in Table~\ref{table:plaalignpartial}. As can be seen in this table, the focused and iterative search method led to detecting all known objects in the shelf image. Hence, the alignment result became the same as the ground truth (GT) as expected. Based on the iteration stopping conditions, we did not resemble the unknown objects at the leftmost and rightmost sections of the planogram to an expected object in the reference planogram. In other words, the threshold values have not been relaxed automatically in the focused and iterative search method to provide false results.

	\begin{table}[htpb]
		\centering
		\caption{Alignment result of the reference and detected planograms for the partially compliant shelf.}	\label{table:plaalignpartial}
		\begin{tabular}{lccccccc}
			\toprule
			$\hat{o}_t$	& A		& $o_1$	& $o_2$	& $o_3$	& A	& $o_4$	& $o_5$		\\
			$\hat{q}_t$	& 0		& 3		& 5		& 5		& 0		& 4	& 2			\\
			\midrule
			$\hat{o}_d$			& U		& $o_1$	& $o_2$	& $o_3$	& E		& $o_4$	& U			\\
			$\hat{q}_d$			& 1		& 3		& 6		& 3		& 1		& 3		& 1			\\
			\midrule
			\textbf{Result}			& NM	& MT	& ME	& MI	& NM	& MI	& NM		\\
			\textbf{GT}				& NM	& MT	& ME	& MI	& NM	& MI	& NM		\\
			\bottomrule
		\end{tabular}
	\end{table}

\section{Experiments}\label{section:Experiments}

We test the proposed method in this section. To do so, we benefit from two datasets. We first introduce them. Then, we provide the object detection performance for the proposed method on these datasets. Afterward, we provide experimental results on the planogram compliance control step. Moreover, we provide sample results of the proposed method.

\subsection{Datasets used in Experiments}

There are several datasets available in literature for similar problems as object detection in retail stores, densely packed object detection, and bounding box extraction. However, they are not suitable for planogram compliance control for the following reasons. First, they do not have the category and annotation information for the objects in the database. Second, they do not provide the reference planogram for the shelves in the database. Therefore, we are left with the Grozi - grocery dataset provided by George and Floerkemeier~\cite{George}. Tonioni and Di Stefano~\cite{Tonioni} manually annotated a subset of the images (consisting of 922 objects) in the Grozi dataset with item-specific bounding boxes. We use these annotations in evaluating our method. Moreover, we were able to compare our object detection performance with theirs.

We also formed our own dataset from the conventional grocery store Migros, Turkey and conducted experiments using it. Images in this dataset are acquired by a Samsung S10 Plus mobile phone with a 16 MP ultra-wide camera. Our dataset is composed of five distinct subsets. The first subset is composed of 14 shelf images to test the performance of the proposed method. We crop these images such that each has only one row of objects. Hence, the total number of images became 28 with 312 objects in them. We also obtain a high resolution image of each object from the web to form the reference planogram. Then, we annotate all the objects in the acquired shelf images. The second subset is composed of 14 shelf images with 124 objects in them. Among these, 64 objects are occluded. Therefore, this subset is specific to test the performance of the proposed method on occluded objects on the shelf. The third subset is composed of 15 images, acquired from five shelves with three different viewing angles, with 75 objects in them. Hence, this subset is specific to test the performance of the proposed method on different viewing angles for the shelf. The fourth subset is composed of 48 objects placed on top of each other. Therefore, it is specific to test the performance of the proposed method on the objects placed on top of each other. We call the combination of all these distinct subsets as the Migros dataset in the following sections. We will publicize this dataset in a Github repository if the manuscript is accepted for publication. Hence, other researchers can test and compare their methods with this study on the same basis.

\subsection{Object Detection Performance: The General Case}

We focus on object detection performance of the proposed method in this section. Here, we assess the object detection as positive when the IoU between the detected and ground-truth bounding boxes is greater than 25\%. We mark the result as true if the detected and ground-truth objects are the same. Otherwise, we mark it as false. We also count undetected bounding boxes of ground-truth as false negatives. Then, we compute the precision, recall, and F1 score as our performance metrics.

We provide object detection performance of the proposed method on the first subset of the Migros dataset without the focused and iterative search step as a baseline. We provide the obtained results in Table~\ref{table:pla_obj_perf_test_single_migros}. As can be seen in this table, the proposed method has a precision higher than 0.9 for all feature extraction methods considered. Therefore, true detection of the correct objects is higher than 90\% and false positive rates are low. The main reason for this result is that we search specific objects in the given shelf image. Hence, it can be detected with high precision if present in the image. All feature extraction methods considered in this study provide recall rates around or higher than 0.9. This indicates that if an object is present in the shelf image, then it can be detected. The missing objects are either occluded or have unexpected face views. The main reason for not detecting such images is that we only have one reference image per object type in our planogram. Finally, the F1 score for all the considered feature extraction methods is around 0.90. This is a high value. Unfortunately, it was not possible to compare this value with manual object detection by a store employee. However, we believe that such a score could not be obtained since the employee should spent time and focus on checking the correctness of the objects on the shelf. Therefore, we expect a high recall value and low precision. Hence, the F1 score will be low.

\begin{table}[htbp]
	\centering
	\caption{Object detection performance of the proposed method without the focused and iterative search step - the first subset of the Migros dataset.}	\label{table:pla_obj_perf_test_single_migros}
	\begin{tabular}{lccc}
		\toprule
		\textbf{Feature Extraction}				&                              	&                       	& 	\\
        \textbf{Method}				                    & \textbf{Precision}	& \textbf{Recall}	& \textbf{F1 score}	\\
		\midrule	
		\textbf{SIFT}		    & 0.958					& 0.920				& 0.939				\\
		\textbf{SURF}		& 0.928					& 0.925				& 0.926				\\
		\textbf{ORB}		& 0.906					& 0.806				& 0.853				\\
		\textbf{AKAZE}		& 0.958					& 0.913				& 0.935				\\
		\textbf{BRISK}		& 0.948					& 0.860				& 0.902				\\
		\bottomrule
	\end{tabular}
\end{table}

We next provide object detection performance of the proposed method on the first subset of the Migros dataset with the focused and iterative search step. We provide the obtained results in Table~\ref{table:pla_obj_perf_test_migros}. As can be seen in this table, compared to the results of baseline object detection method given in Table~\ref{table:pla_obj_perf_test_single_migros}, our proposed focused and iterative search method achieves higher object detection performance independent of the feature extraction method considered. To be more specific, decreasing the threshold values iteratively leads to higher recall rates. Besides, the focused and iterative search method also decreases false detections in the previous steps. This increases precision values. Therefore, we can reach F1 scores around 0.98. This is a remarkable result for the given problem.

\begin{table}[htbp]
	\centering
	\caption{Object detection performance of the proposed method with the focused and iterative search step - the first subset of the Migros dataset.} \label{table:pla_obj_perf_test_migros}
	\begin{tabular}{lccc}
		\toprule
		\textbf{Feature Extraction}				&                              	&                       	& 	\\
        \textbf{Method}				                    & \textbf{Precision}	& \textbf{Recall}	& \textbf{F1 score}	\\
		\midrule
		\textbf{SIFT}		    & 0.997					& 0.987				& 0.992				\\
		\textbf{SURF}		& 0.984					& 0.990				& 0.987				\\
		\textbf{ORB}		& 0.961					& 0.980				& 0.970				\\
		\textbf{AKAZE}		& 0.994					& 0.984				& 0.989				\\
		\textbf{BRISK}		& 0.987					& 0.974				& 0.981				\\
		\bottomrule
	\end{tabular}
\end{table}

\begin{figure}[htbp]
	\centering	
	\includegraphics[width=0.9\columnwidth]{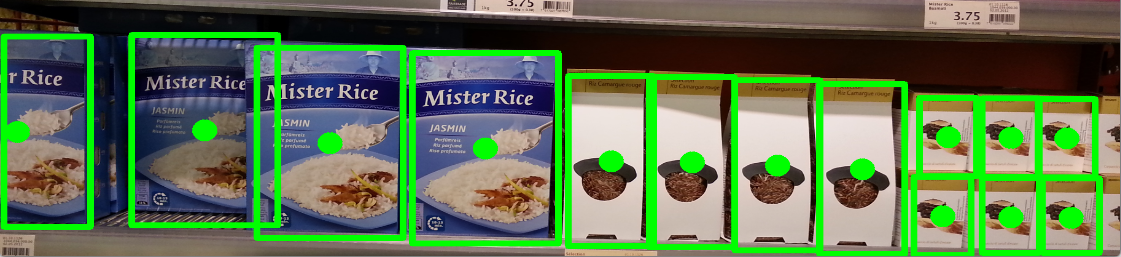}
	\caption{Bounding box of the detected objects from the representative shelf image in the Grozi dataset with focused and iterative search.}\label{fig:testimages_result_grozi}
\end{figure}

We also tested the proposed object detection method with the focused and iterative search on the Grozi dataset. We provide a representative image from the dataset and extracted bounding box for the detected objects, when SIFT is used as the local feature extraction method, in Fig.~\ref{fig:testimages_result_grozi}. In this image, one object is seen in half form. Another object is placed slightly backward compared to others. There are also objects placed on top of each other. As can be seen in Fig.~\ref{fig:testimages_result_grozi}, all the objects are detected correctly in the representative image with the proposed method.

We provide object detection results with the focused and iterative search step on the Grozi dataset in Table~\ref{table:pla_obj_perf_test_grozi}. As can be seen in this table, F1 scores obtained by different local feature extraction methods decrease compared to the Migros dataset. Lower F1 scores in the Grozi dataset are because of two main reasons. First, images in this dataset are blurry and have low resolution. Second, the proposed method is sensitive to the shelf height and full visibility of the shelf. There are some images in the Grozi dataset which violate this constraint. Therefore, they led to low performance. Even with these negative effects, the proposed method was able to produce F1 scores around 0.96.

\begin{table}[htbp]
	\centering
	\caption{Object detection performance of the proposed method with the focused and iterative search step - the Grozi dataset.}
	\label{table:pla_obj_perf_test_grozi}
	\begin{tabular}{lccc}
		\toprule
		\textbf{Feature Extraction}				&                              	&                       	& 	\\
        \textbf{Method}				                    & \textbf{Precision}	& \textbf{Recall}	& \textbf{F1 score}	\\
		\midrule
		\textbf{SIFT}		    & 0.953					& 0.981				& 0.967			\\
		\textbf{SURF}		& 0.928					& 0.981				& 0.954			\\
		\textbf{ORB}		& 0.887					& 0.966				& 0.925			\\
		\textbf{AKAZE}		& 0.955					& 0.966				& 0.961			\\
		\textbf{BRISK}		& 0.946					& 0.974				& 0.960			\\
		\bottomrule
	\end{tabular}
\end{table}

We also compared the proposed method with the results obtained by Tonioni and Di Stefano~\cite{Tonioni}. To do so, we used the same settings suggested by them such as setting the IoU as 50\%. Tonioni and Di Stefano reported an F1 score as 0.904 for their object detection method which uses BRISK and best-buddies similarity method. We obtained the F1 scores as 0.922, 0.910, 0.869, 0.917, and 0.912 for the SIFT, SURF, ORB, AKAZE, and BRISK, respectively. As can be seen in these results, even by choosing their settings, the proposed method was able to provide higher or same F1 scores on four of the five local feature extraction methods.

\subsection{Object Detection Performance: Occluded Objects}

\begin{figure}[htbp]
	\centering	
	\includegraphics[width=0.9\columnwidth]{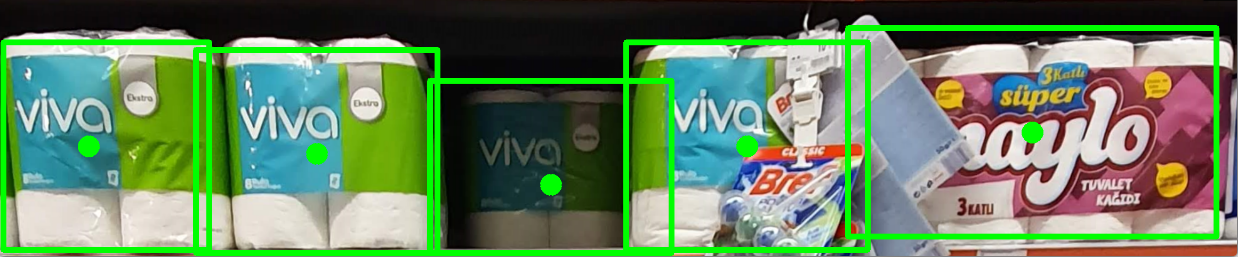}
	\caption{Bounding box of the detected objects from the representative shelf image from the second subset of the Migros dataset.}\label{fig:testimages_result_occlusion}
\end{figure}

We next consider the effect of occlusion on the object detection performance of the proposed method. To do so, we use the second subset of the Migros dataset. We provide a representative image from this subset and corresponding object detection results, when SIFT is used as the local feature extraction method, in Fig.~\ref{fig:testimages_result_occlusion}. As can be seen in this figure, our proposed method can detect occluded objects in the image.

We tabulate object detection results of the proposed method on the second subset of the Migros dataset in Table~\ref{table:pla_obj_perf_test_occlusion}. As can be seen in this table, ORB and AKAZE has higher F1 scores compared to other local feature extraction methods. The possible reason for this performance is as follows. ORB and AKAZE produce more keypoints compared to other methods. Hence, the object can still be detected even if it is occluded. SURF and BRISK have similar and average results. SIFT performs below average compared to other methods. The main reason for this low performance is based on the low number of extracted keypoints from the images via SIFT.

\begin{table}[htbp]
	\centering
	\caption{Object detection performance of the proposed method - the second subset of the Migros dataset.}\label{table:pla_obj_perf_test_occlusion}
	\begin{tabular}{lcccl}
		\toprule
		\textbf{Feature Extraction}				&                              	&                       	& 	\\
        \textbf{Method}				                    & \textbf{Precision}	& \textbf{Recall}	& \textbf{F1 score}	\\
		\midrule
		\textbf{SIFT}		    & 0.979					& 0.730				& 0.836			\\
		\textbf{SURF}		& 0.864					& 0.966				& 0.912			\\
		\textbf{ORB}		& 0.906					& 0.951				& 0.928			\\
		\textbf{AKAZE}		& 0.965					& 0.902				& 0.932			\\
		\textbf{BRISK}		& 0.931					& 0.871				& 0.900			\\
		\bottomrule
	\end{tabular}
\end{table}

\subsection{Object Detection Performance: The Effect of Viewing Angle}

\begin{figure}[htbp]
	\centering
	\subfigure[Top.]{\includegraphics[width=0.9\columnwidth]{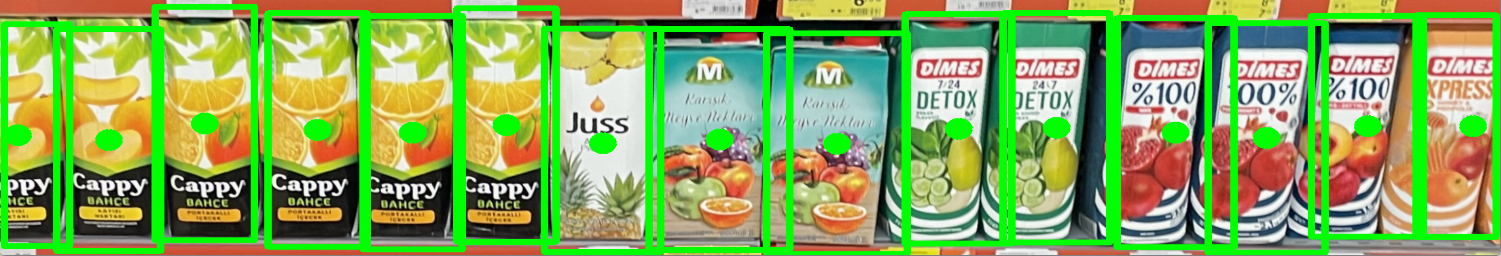}\label{fig:res_local_sift_top}}\\
	\subfigure[Middle.]{\includegraphics[width=0.9\columnwidth]{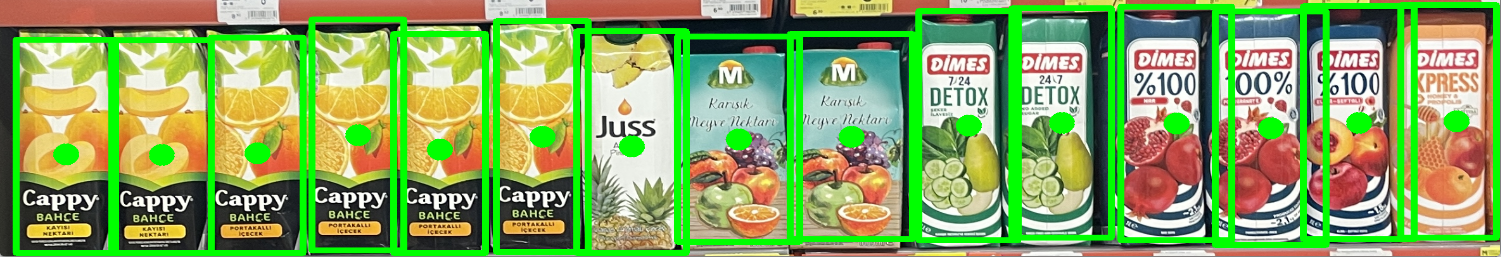}\label{fig:res_local_sift_middle}}\\
	\subfigure[Bottom.]{\includegraphics[width=0.9\columnwidth]{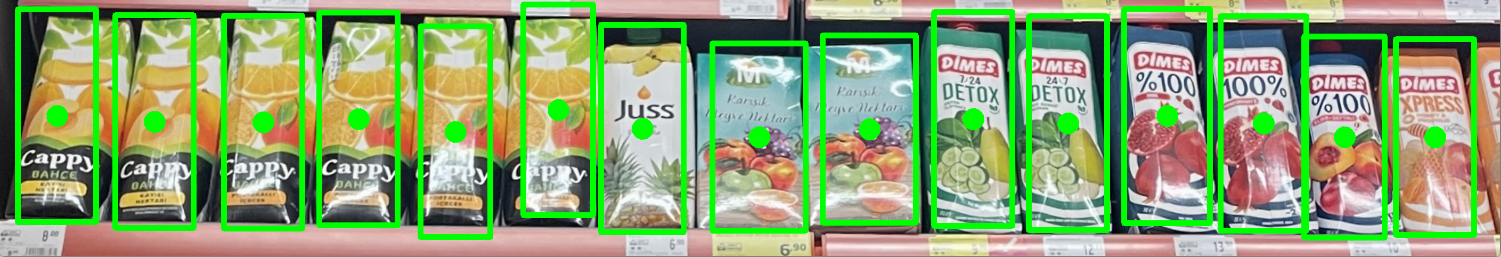}\label{fig:res_local_sift_bottom}}
	\caption{Bounding box of the detected objects from the representative shelf images from the third subset of the Migros dataset.}\label{fig:testimages_result_camangle}
\end{figure}

We next test the proposed method using shelf images from different viewing angles. To do so, we use the third subset of the Migros dataset. We provide representative images from this subset and corresponding object detection results, when SIFT is used as the local feature extraction method, in Fig.~\ref{fig:testimages_result_camangle}. As can be seen in this figure, the proposed method can detect objects with different camera placements.

We tabulate object detection results of the proposed method, in terms of F1 score, on the third subset of the Migros dataset in Table~\ref{table:pla_obj_perf_test_camangle}. As can be seen in this table, the camera placement to the top and middle shelves lead to higher F1 score values compared to the bottom shelf. This result is expected since as we set the camera to the bottom shelf, most objects will not be visible completely. Therefore, the proposed method requires placing the camera to the middle or top location.

\begin{table}[htbp]
	\centering
	\caption{Object detection performance, in terms of F1 score, of the proposed method - the third subset of the Migros dataset.}\label{table:pla_obj_perf_test_camangle}
	\begin{tabular}{lccc}
		\toprule
		\textbf{Feature Extraction}				&   \multicolumn{3}{c}{\textbf{Shelf Location}} 	\\
        \textbf{Method}				                    & \textbf{Top}	& \textbf{Middle}	& \textbf{Bottom}	\\
		\midrule
		\textbf{SIFT}		    & 0.944				& 0.959				& 0.864				\\
		\textbf{SURF}		& 0.966				& 0.960				& 0.857				\\
		\textbf{ORB}		& 0.897				& 0.952				& 0.782				\\
		\textbf{AKAZE}		& 0.929				& 0.952				& 0.872				\\
		\textbf{BRISK}		& 0.913				& 0.916				& 0.837				\\
		\bottomrule
	\end{tabular}
\end{table}

\subsection{Object Detection Performance: Objects on Top of Each Other}

\begin{figure}[htbp]
	\centering	
	\includegraphics[width=0.9\columnwidth]{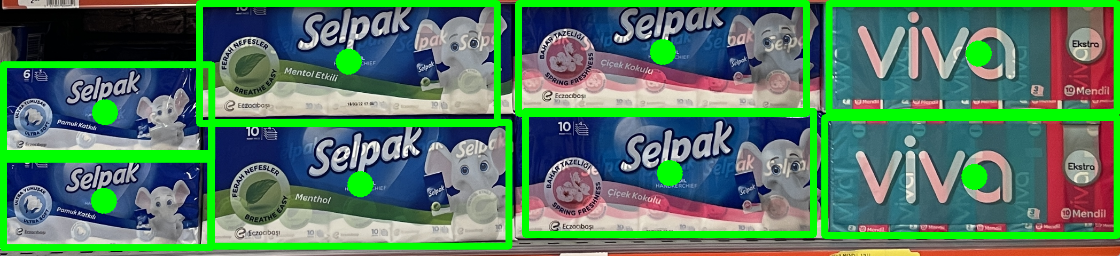}
	\caption{Bounding box of the detected objects from the representative shelf image from the fourth subset of the Migros dataset.}\label{fig:testimages_result_multirow}
\end{figure}

We finally consider performance of the proposed method when objects are placed on top of each other on the shelf. To do so, we use the fourth subset of the Migros dataset. We provide a representative image from this subset and corresponding object detection results, when SIFT is used as the local feature extraction method, in Fig.~\ref{fig:testimages_result_multirow}. As can be seen in this figure, the proposed method can detect objects on top of each other.

We tabulate object detection results of the proposed method on the fourth subset of the Migros dataset in Table~\ref{table:pla_obj_perf_test_multirow}. As can be seen in this table, all the feature extraction methods, except SIFT, provide fairly good F1 score values. The low performance in SIFT is because of extra false positives. Besides, the fourth subset of the Migros dataset has limited samples. This also leads to the low performance. On the other hand, placement of the objects on top of each other require strict organization. Therefore, feature extraction methods generally work fairly well on them.

\begin{table}[htbp]
	\centering
	\caption{Object detection performance of the proposed method - the fourth subset of the Migros dataset.}\label{table:pla_obj_perf_test_multirow}
	\begin{tabular}{lcccl}
		\toprule
		\textbf{Feature Extraction}				&                              	&                       	& 	\\
        \textbf{Method}				                    & \textbf{Precision}	& \textbf{Recall}	& \textbf{F1 score}	\\
		\midrule
		\textbf{SIFT}		    & 0.827					& 1.000				& 0.905				\\
		\textbf{SURF}		& 0.979					& 1.000				& 0.990				\\
		\textbf{ORB}		& 0.959					& 1.000				& 0.979				\\
		\textbf{AKAZE}		& 0.959					& 1.000				& 0.979				\\
		\textbf{BRISK}		& 0.923					& 1.000				& 0.960				\\
		\bottomrule
	\end{tabular}
\end{table}

\subsection{Planogram Compliance Control Performance}

In this section, we consider performance of the proposed method on the planogram compliance control step. To do so, we benefit from the first subset of the Migros dataset and Grozi dataset. For the Migros dataset, we asked the responsible department from Migros to generate the corresponding reference planogram for the shelf images considered. For the Grozi dataset, we benefit from the method proposed by Tonioni and Di Stefano~\cite{Tonioni}. Hence, we converted their graph based representation to our format as ground truth. Then, we provide the planogram compliance control performance on the dataset.

There is no consensus on measuring the planogram compliance control performance in literature. The closest study for this purpose is by Liu~\emph{et al.}~\cite{Liu}. Unfortunately, this study focuses on a graph theory based representation. Therefore, we provide our own measures based on precision, recall, and F1 score values. For the planogram compliance control, we calculate these values as follows. We assess the alignment result as true positive if it is the same as ground truth for an object group. Otherwise, we mark it as false positive. We assess the result as false positive if we find an alignment result for an object group that is not in ground truth. We assess the alignment result as false negative if there is an object group in ground truth but not in the alignment result.

We provide planogram compliance control performance results for the first subset of the Migros dataset in Table~\ref{table:pla_compliance_perf_test_migros}. As can be seen in this table, SIFT and AKAZE have higher F1 scores compared to other local feature extraction methods. SURF and BRISK have similar and average results. ORB performs below average compared to other methods. We can also consider Table~\ref{table:pla_compliance_perf_test_migros} from a general perspective. The most important observation here is that the planogram compliance control performance is directly related to the object detection performance given in Table~\ref{table:pla_obj_perf_test_migros}. This can be expected since correctly detecting objects in the shelf image directly affects the corresponding planogram formation from the shelf.

\begin{table}[htbp]
	\centering
	\caption{Planogram compliance control performance of the proposed method - the first subset of the Migros dataset.}
	\label{table:pla_compliance_perf_test_migros}
	\begin{tabular}{lcccl}
		\toprule
		\textbf{Feature Extraction}				&                              	&                       	& 	\\
        \textbf{Method}		& \textbf{Precision}	& \textbf{Recall}	& \textbf{F1 score}	\\
		\midrule
		\textbf{SIFT}		& 0.954					& 0.980				& 0.966			\\
		\textbf{SURF}		& 0.929					& 0.986				& 0.957			\\
		\textbf{ORB}		& 0.873					& 0.979				& 0.923			\\
		\textbf{AKAZE}		& 0.961					& 0.987				& 0.973			\\
		\textbf{BRISK}    	& 0.916					& 0.979				& 0.947			\\
		\bottomrule
	\end{tabular}
\end{table}

We provide planogram compliance control performance results for the Grozi dataset in Table~\ref{table:pla_compliance_perf_test_grozi}. As can be seen in this table, the performance results tabulated are lower than the Migros dataset. This is because of the low object detection performance of the proposed method on this dataset.

\begin{table}[htbp]
	\centering
	\caption{Planogram compliance control performance of the proposed method - the Grozi dataset.}
	\label{table:pla_compliance_perf_test_grozi}
	\begin{tabular}{lcccl}
		\toprule
		\textbf{Feature Extraction}				&                              	&                       	& 	\\
        \textbf{Method}		& \textbf{Precision}	& \textbf{Recall}	& \textbf{F1 score}	\\
		\midrule
		\textbf{SIFT}		& 0.846					& 0.986				& 0.911			\\
		\textbf{SURF}		& 0.880					& 0.998				& 0.935			\\
		\textbf{ORB}		& 0.806					& 0.988				& 0.888			\\
		\textbf{AKAZE}		& 0.876					& 0.993				& 0.931			\\
		\textbf{BRISK}    	& 0.836					& 0.991				& 0.907			\\
		\bottomrule
	\end{tabular}
\end{table}

We can summarize the key findings from Tables \ref{table:pla_compliance_perf_test_migros} and \ref{table:pla_compliance_perf_test_grozi} for the Migros and Grozi datasets as follows. The precision values are lower than recall in both tables. This indicates that we have less number of false negatives. In other words, the proposed method was able to detect most of the ground truth objects in the given image. Unfortunately, we have high false positives ar the same time. Therefore, we have relatively low precision values in both tables.

\section{Final Comments}

Planogram compliance control is an important problem in retail sector. In this study, we propose a solution to this problem via pattern recognition and computer vision tools. Our aim is to detect objects in shelf images and check their placement wrt the corresponding reference planogram at hand. This problem is slightly different from the usual object detection applications. The main reason is that, we have the reference planogram such that we know where to find a target object. However, this information may be faulty due to several reasons as follows. There may be missing or extra objects on the shelf different from the ones in the planogram. Some objects may be occluded or placed backward on the shelf. Moreover, planogram compliance control requires relative placement of the objects as well as their number on the shelf. Therefore, we proposed a novel method different from the available ones in literature. Our novelty comes in three different steps. First, we modified and enhanced the ISM for object detection and boundary extraction from the shelf image. Second, we modified the NW algorithm for planogram compliance control. Third, we introduced an iterative and focused search to improve the object detection and planogram compliance control steps. Therefore, the proposed method is based on object detection, alignment, and focused iterative search. We tested these three steps on different datasets and tabulated their performance. While doing so, we picked challenging scenarios as occlusion, different viewing angles, and objects placed on top of each other. We also compared the proposed method with a similar one in literature on the same dataset. We discussed the obtained results to indicate the strengths and weaknesses of our method. Based on these, we can claim that the proposed method can be used in retail sector and smart retail stores with confidence. The next step for us is extending our method to run on low power embedded systems. Hence, it can be used in stand alone form in retail sector.

\section*{Acknowledgement}

 This work is supported by TUBITAK under project no 5190042.

\bibliographystyle{elsarticle-num}
\bibliography{planogram_control}

\end{document}